\documentclass[sigconf]{acmart}
\AtBeginDocument{%
  }

\copyrightyear{2026}
\acmYear{2026}
\setcopyright{none}          % [arXiv author version]
\acmConference[MM '26]{Proceedings of the 34th ACM International Conference on Multimedia}{November 10--14, 2026}{Rio de Janeiro, Brazil}
\acmBooktitle{Proceedings of the 34th ACM International Conference on Multimedia (MM '26), November 10--14, 2026, Rio de Janeiro, Brazil}
\acmDOI{10.1145/3767308.3837680}
\acmISBN{979-8-4007-2213-4/2026/11}

\begin{document}

\title{SoftRerank: Hierarchical Soft Fusion with Candidate-Label Reranking for Long-Tailed Micro-Action Recognition}

\author{Yichi Zhang}
\authornote{Both authors contributed equally to this research.}
\affiliation{%
  \institution{University of Science and Technology of China}
  \city{Hefei}
  \country{China}
}
\email{charleszhang@mail.ustc.edu.cn}

\author{Zhichao Xia}
\authornotemark[1]
\affiliation{%
  \institution{University of Science and Technology of China}
  \city{Hefei}
  \country{China}
}
\email{xiazc118@mail.ustc.edu.cn}

\author{Yanjun Chi}
\affiliation{%
  \institution{University of Science and Technology of China}
  \city{Hefei}
  \country{China}
}
\email{yjChi@mail.ustc.edu.cn}

\author{Lingsi Zhu}
\affiliation{%
  \institution{University of Science and Technology of China}
  \city{Hefei}
  \country{China}}
\email{ls-zhu24@mail.ustc.edu.cn}

\author{Yuefeng Zou}
\affiliation{%
  \institution{University of Science and Technology of China}
  \city{Hefei}
  \country{China}}
\email{zouyuefeng.00@mail.ustc.edu.cn}

\author{Jun Yu}
\authornote{Corresponding author.}
\affiliation{%
  \institution{University of Science and Technology of China}
  \city{Hefei}
  \country{China}}
\email{harryjun@ustc.edu.cn}

\author{Qingsong Liu}
\affiliation{%
  \institution{Unisound AI Technology Co., Ltd.}
  \city{Beijing}
  \country{China}}
\email{liuqingsong@unisound.com}

\author{Jianqing Sun}
\affiliation{%
  \institution{Unisound AI Technology Co., Ltd.}
  \city{Beijing}
  \country{China}}
\email{sunjianqing@unisound.com}

\author{Shengping Liu}
\affiliation{%
  \institution{Unisound AI Technology Co., Ltd.}
  \city{Beijing}
  \country{China}}
\email{liushengping@unisound.com}

\renewcommand{\shortauthors}{Yichi Zhang et al.}

\begin{abstract}
Micro-actions are subtle, low-intensity non-verbal behaviors that provide cues to fine-grained human states, including emotions and intentions. Recognizing them remains difficult because they are brief, contain weak visual changes, and often exhibit similar motion patterns across categories. This paper addresses these challenges with a fine-grained micro-action recognition method that combines full fine-tuning of InternVideo2.5, hierarchical soft fusion, and a lightweight candidate-label reranker. For the long-tailed label distribution in MA-52, we use class-balanced sampling and inverse-frequency reweighting to reduce the effect of frequent classes during training. We fine-tune InternVideo2.5 end to end and attach coarse and group-conditional fine-grained classification heads to the shared video representation, improving the consistency between coarse and fine predictions. For ambiguous samples, the candidate-label reranker uses hard samples and video-label matching to focus on easily confused fine-grained actions. Experiments validate the proposed method, which achieves a 79.99\% F1-mean on MA-52 and ranks first in the 3rd Micro-Action Analysis Grand Challenge at ACM Multimedia 2026.
\end{abstract}



\maketitle

% ---- arXiv author-version notice ----
\renewcommand{\thefootnote}{}
\footnotetext{Accepted to the 34th ACM International Conference on Multimedia (MM '26), November 10--14, 2026, Rio de Janeiro, Brazil. This is the author's version of the work, posted here for personal use. The definitive Version of Record is available at \url{https://doi.org/10.1145/3767308.3837680}.}
\setcounter{footnote}{0}
\renewcommand{\thefootnote}{\arabic{footnote}}
%% ---- end notice ----

\section{Introduction}
Action recognition is a fundamental problem in computer vision, with applications in intelligent surveillance, medical assistance, human-computer interaction, and behavior analysis \cite{guo2024mac,li2025mac,li2026mac}. Micro-Action Recognition (MAR) \cite{gu2025motion,guo2026rethinking} focuses on subtle, short-duration movements, such as slight head motions, small hand gestures, and minor posture changes, which provide behavioral and psychological cues for emotion understanding, psychological assessment, and safety monitoring \cite{guo2024benchmarking,li2023joint,li2025mmad}. Unlike conventional actions, micro-actions exhibit weak motion intensity, sparse temporal evidence, and small inter-class variations, requiring models to capture brief, localized motion patterns while distinguishing visually similar categories. MA-52 is a representative MAR benchmark containing 52 fine-grained categories organized into 7 coarse-grained body-motion groups, with each fine-grained label assigned to a unique parent group \cite{li2025mmad,liu2026self}. This hierarchy reflects the natural organization of micro-actions and supports evaluation at both body-motion and fine-action levels, but also requires separating actions across groups and distinguishing similar actions within each group. For example, turning the head, nodding, shaking the head, tilting the head, and raising the head share similar spatial regions and motion trajectories, making their discrimination dependent on subtle differences in motion direction, temporal evolution, and visual context.

Recognition on MA-52 involves three closely related challenges. First, the long-tailed fine-grained label distribution allows frequent classes to dominate optimization and limits the representation quality of rare classes, particularly within body-motion groups where visually similar categories compete under substantially different sample frequencies. Second, discriminative evidence may occur in only a few frames and reflect minor appearance or motion changes, requiring precise spatiotemporal representation learning. Third, although the coarse-to-fine hierarchy provides useful structural information, incorporating it is non-trivial. A flat 52-way classifier ignores label-space organization, whereas a hard coarse-to-fine cascade restricts fine-grained prediction to one coarse group and may propagate an incorrect coarse prediction to the final decision. Moreover, even a strong fine-tuned model can produce unstable top-1 predictions when several semantically related labels receive similar probabilities.

To address these challenges, we propose a unified MAR framework based on full end-to-end fine-tuning of the InternVideo2.5 backbone \cite{wang2025internvideo}. Its long-context and fine-grained spatiotemporal modeling capacity enables the shared video representation to adapt to subtle motion differences without an additional temporal contextualization module. To reduce optimization bias under the long-tailed distribution, training incorporates class-balanced sampling, inverse-frequency loss reweighting within each parent group, stronger augmentation for rare classes, and hierarchical Mixup, which interpolates coarse- and fine-level supervision according to the label hierarchy to maintain semantic consistency during data mixing. On the shared representation, hierarchical soft fusion performs hierarchy-consistent global prediction. A coarse head estimates probabilities over the 7 body-motion groups, while a group-conditional fine-grained head models fine-grained categories within each group. Each fine-grained probability is factorized into its coarse-group and group-conditional probabilities. Unlike hard cascading, the final decision globally searches all 52 fine-grained categories, preserving hierarchical consistency without excluding labels outside the coarse top-1 group. This formulation uses coarse information to regularize fine-grained prediction while reducing error propagation from incorrect coarse decisions.

We further introduce a lightweight candidate-label reranker as a selective correction stage for ambiguous predictions. The reranker is trained using out-of-fold predictions, which provide hard samples from models that have not observed the corresponding training samples. For each hard sample, a compact candidate set is constructed from high-scoring incorrect predictions, labels within the same body-motion group, and frequently confused labels identified from confusion statistics. During inference, reranking is activated only for low-confidence predictions or those with a small margin between the two highest probabilities. The reranker compares the video representation with candidate-label representations and refines the decision within this restricted set. Together, hierarchical soft fusion and selective candidate reranking form a two-stage inference process that first performs hierarchy-consistent global prediction and then corrects residual ambiguities among a small set of confusing fine-grained actions.

Our contributions are as follows:
\begin{itemize}

\item[$\bullet$] We propose an end-to-end fine-tuned micro-action recognition framework based on InternVideo2.5, where imbalance-aware sampling, inverse-frequency reweighting, and rare-class augmentation are jointly used to improve representation learning on the long-tailed MA-52 dataset.

\item[$\bullet$] We introduce a hierarchical soft fusion strategy to jointly model coarse body-motion groups and fine-grained action labels, maintaining coarse--fine consistency while reducing error propagation from hard coarse-to-fine decisions.

\item[$\bullet$] We design a lightweight candidate-label reranker trained with out-of-fold hard samples and confusion-aware candidate sets, which refines ambiguous predictions by comparing video features with a small set of likely confused action labels under limited additional computation.

\item[$\bullet$] We achieve an F1-mean of 79.99\% on MA-52, ranking first in the 3rd Micro-Action Analysis Grand Challenge at ACM Multimedia 2026.
\end{itemize}

\begin{figure}[ht]
    \includegraphics[width=\linewidth, height=\textheight, keepaspectratio]{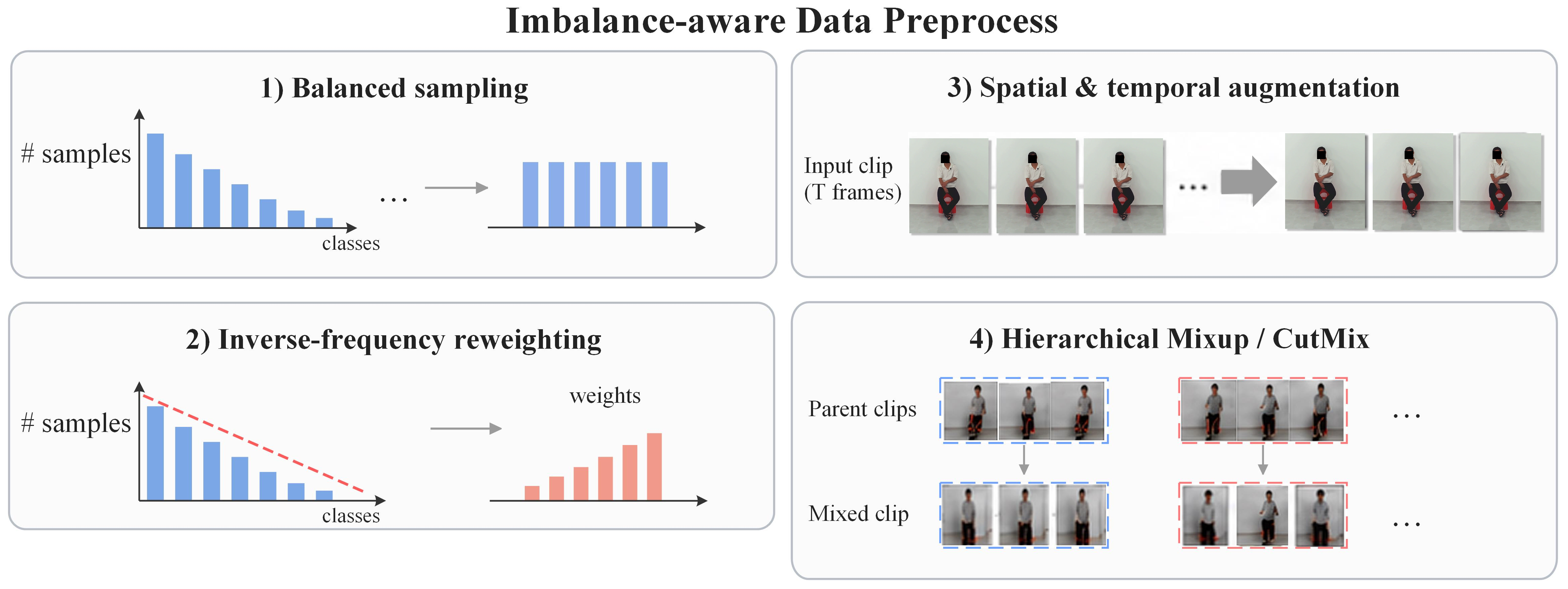}
    \caption{Overview of MA-52 and the imbalance-aware data preprocessing strategies, including class-balanced sampling, inverse-frequency reweighting, and stronger augmentation for rare classes.}
    \Description{Two-panel overview of the MA-52 dataset. The left panel plots the long-tailed sample counts across the 52 fine-grained micro-action classes grouped into 7 body-level categories. The right panel illustrates the imbalance-aware preprocessing pipeline, showing class-balanced sampling and inverse-frequency reweighting applied within each parent group.}
    \label{fig:data-preprocessing}
\end{figure}

\begin{figure*}[ht]
    \includegraphics[width=\linewidth, height=\textheight, keepaspectratio]{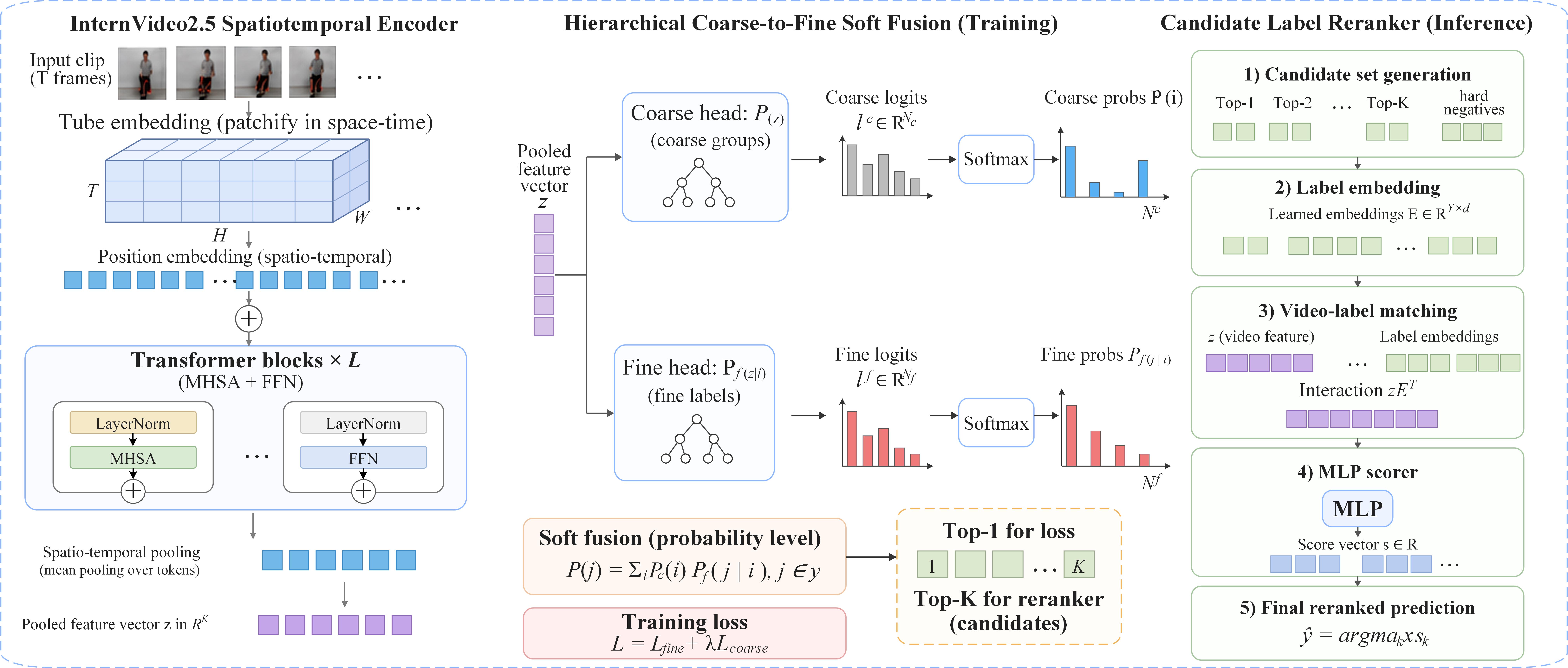}
    \caption{Overview of the proposed hierarchical micro-action recognition framework, which integrates imbalance-aware preprocessing, InternVideo2.5-based spatiotemporal representation learning, hierarchical coarse-to-fine soft fusion, and candidate-label reranking for accurate fine-grained micro-action recognition.}
    \Description{Block diagram of the proposed hierarchical micro-action recognition framework. Video clips pass through imbalance-aware preprocessing into a fully fine-tuned InternVideo2.5 backbone, which feeds a shared representation to a coarse body-level head and a group-conditional fine-grained head. The two head outputs are combined by hierarchical soft fusion, and ambiguous samples are then refined by a lightweight candidate-label reranker that scores video-label matching.}
    \label{fig:framework}
\end{figure*}

\section{Related Work}

\textbf{Skeleton- and Pose-Based Methods.}
These methods represent the body as a graph of joints and bones and feed pose sequences
(keypoints) extracted from video into the model. The design effort goes into graph neural
networks or spatiotemporal models that capture both the spatial relations between joints
and their motion over time. Early SkelAct work \cite{li2017skeleton,li2018co-occurrence,ye2020dynamic,xu2022topology}
combined skeletal cues with RGB appearance, while toolkits such as PYSKL
\cite{duan2022pysklgoodpracticesskeleton} and OpenPose \cite{caoOpenPoseRealtimeMultiPerson2019}
made pose extraction practical. Recent methods include BlockGCN \cite{zhouBlockGCNRedefineTopology2024}, ProtoGCN
\cite{liuRevealingKeyDetails2025}, DeGCN \cite{myungDeGCNDeformableGraph2024}, and
Hyperformer \cite{zhouHypergraphTransformerSkeletonbased2023}, which models higher-order
joint dependencies using hypergraph self-attention. SeFAR
\cite{huangSeFARSemisupervisedFinegrained2025a} targets semi-supervised recognition,
whereas DSTA-Net \cite{shi2020dstanet} decouples spatial and temporal attention. Skeletal input is robust to background, lighting, and appearance
shifts and is cheap to process, but it depends on pose-estimation quality and struggles
under occlusion, fast motion, and exactly the fine appearance details that micro-actions
hinge on.

\textbf{Multimodal Integration Methods.}
To get past the limits of any single modality, these methods fuse complementary signals into
a more robust action representation. Our framework draws on contrastive vision-language work:
CLIP \cite{radford2021learningtransferablevisualmodels}, M2-CLIP \cite{wangM2CLIPMultimodalMultitask2024},
and TC-CLIP \cite{kimLeveragingTemporalContextualization2024a}, the first emphasizing intra
and inter-modal alignment and the latter focusing on temporal alignment. MMCL-Action
\cite{liu2024mmcl} similarly relies on multimodal contrastive learning. ExACT
\cite{zhouExACTLanguageguidedConceptual2024} and the DailyDVS-200 benchmark
\cite{wangDailyDVS200ComprehensiveBenchmark2024} explore event-camera input, and MM-CDFSL
\cite{hatanoMultimodalCrossDomainFewShot2025} studies cross-domain few-shot learning across
RGB, flow, and audio. DualActNet \cite{yuDualActNetExploitingSlowFast2024} uses a SlowFast
design to fuse information at different temporal rates. For MAR specifically, MANet
\cite{guo2024benchmarking} outperformed nine competing methods, and PCAN
\cite{li2025prototypical,chen2024prototype} reduces the ambiguity between visually similar
micro-actions. More recently, multimodal fusion that combines body-level context with
fine motion cues has become a common recipe for MA-52, with strong systems pairing a
body-level backbone and a motion-level backbone rather than relying on one model alone
% TODO: 补一篇能支撑"双骨干/多模态融合是当前 MA-52 主流方案"这一趋势断言的文献
% （如有 leaderboard 对比的综述或参赛报告）；若仅作为本文设计动机，可把此句软化后删掉 \cite{}。
\cite{gu2025mm}.

\textbf{Video Foundation Models and Efficient Supervision.}
Multimodal Large Language Models (MLLMs) have reshaped video understanding and, with it,
action recognition. MA-FSAR \cite{xing2024mafsarmultimodaladaptationclip} adds temporal
awareness to CLIP through a Global Timing Adapter and Text-guided Prototyping under a
PEFT budget, while MotionSight \cite{du2025motionsightboostingfinegrainedmotion} uses a
zero-shot "visual spotlight" that pairs trajectory tracking with motion-blur enhancement to
sharpen perception of subtle object and camera motion. AAPL \cite{yoshida2025aapl} reduces annotation effort through action-agnostic frame
sampling and point-level supervision while remaining competitive across temporal action
detection benchmarks.
InternVideo2.5 \cite{wang2025internvideo} trains progressively, unifying masked video modeling
with next-token prediction to capture structure and semantics at several levels, and STIA
\cite{yu2024end} aggregates spatiotemporal information for end-to-end online feature
learning. The broader trend is toward larger video foundation models and instruction-tuned
video LLMs being adapted for fine-grained motion, where the open question is less about raw
capacity and more about fine-tuning efficiency and countering the long-tail bias inherited
from web-scale pretraining \cite{shang2025cross}.

\section{Methodology}

As shown in Figure~\ref{fig:framework}, our method builds a unified MAR framework with a fully fine-tuned InternVideo2.5 backbone, using imbalance-aware sampling, inverse-frequency reweighting, and rare-class augmentation to improve representation learning under the long-tailed label distribution of MA-52. Based on the shared video representation, hierarchical soft fusion jointly models coarse body-motion groups and fine-grained action labels, and a lightweight candidate-label reranker refines ambiguous predictions by comparing video features with a compact set of likely confused labels.

\subsection{Data Preprocessing}
\label{sec:data}

\textbf{Imbalance-Aware Training.}
MA-52 has a long-tailed fine-grained label distribution, and the hierarchy in
Sec.~\ref{sec:soft_fusion} handles cross-group confusion but not the intra-group
tail. We address the tail on the data side with four strategies.

\emph{Class-balanced sampling.} Let $N_f$ be the sample count of fine-grained class $f$.
We sample from a smoothed distribution
\[
q_f \propto N_f^{\,\tau},\qquad \tau\in[0,1],
\]
where $\tau{=}1$ is instance-balanced and $\tau{=}0$ is class-balanced. An
intermediate $\tau$ keeps head classes from dominating the gradient while
avoiding repeated tail samples.

\emph{Inverse-frequency reweighting.} We weight the fine-grained cross-entropy by
$w_f \propto N_f^{-\gamma}$, $\gamma\in[0,1]$, computed within each parent group.
The group-conditional head then does not collapse onto the most frequent member
of its group.

\emph{Stronger augmentation on rare classes.} We scale augmentation intensity by
class frequency. Tail classes receive wider color perturbation and a larger
motion-amplification factor $\alpha$, which enlarges their effective support
without new data.

\emph{Hierarchical Mixup/CutMix.} For two samples $(x_a,f_a)$ and $(x_b,f_b)$
mixed with coefficient $\lambda$, we interpolate labels at both tree levels,
\[
\tilde{y}^{\text{fine}} = \lambda\,\mathbf{e}_{f_a}+(1-\lambda)\,\mathbf{e}_{f_b},
\qquad
\tilde{y}^{\text{coarse}} = \lambda\,\mathbf{e}_{c(f_a)}+(1-\lambda)\,\mathbf{e}_{c(f_b)}.
\]
The mixed target respects the parent assignment $c(\cdot)$ and supervises both
heads. Cross-group mixing places soft samples on group boundaries, where soft
fusion does its arbitration.

\subsection{Backbone and Full Fine-tuning}
\label{sec:backbone}

We build our model on a single InternVideo2.5 backbone $\Phi(\cdot)$, which
maps an input video $x$ to a pooled representation $z=\Phi(x)$ shared by the
coarse and fine-grained heads of Sec.~\ref{sec:soft_fusion}. Compared with the
InternVideo2 backbone used in our previous solution, InternVideo2.5 provides
stronger long-context and fine-grained spatiotemporal modeling, so the
external Temporal Contextualization module previously needed to inject
cross-frame context is no longer required; the backbone already captures the
temporal cues that micro-actions depend on.

In contrast to the parameter-efficient fine-tuning adopted last year, we fine-tune
the backbone end to end. Micro-actions differ from each other by subtle
appearance and motion details, and a frozen or adapter-only backbone leaves the
shared features only partially aligned with the task. Full fine-tuning lets the
representation adapt to these fine distinctions, and, since both heads read from
the same $z$, it also lets the coarse objective regularize the shared features
in the direction that benefits the fine-grained task, consistent with the joint objective
described in Section~\ref{sec:soft_fusion}.

Full fine-tuning of a billion-scale backbone on a long-tailed dataset of this
size is prone to overfitting, so we pair it with the data-side rebalancing and
augmentation of Sec.~\ref{sec:data} and a small, layer-wise decayed learning
rate on the backbone. Frames are sampled from each clip, center-cropped and
resized to $224\times224$, and encoded by $\Phi(\cdot)$; the two heads are then
attached on top of $z$ as described in Sec.~\ref{sec:soft_fusion}. Training uses
mixed precision and activation checkpointing to keep the memory footprint of
end-to-end tuning tractable.

\subsection{Hierarchical Soft Fusion}
\label{sec:soft_fusion}

Let $c\in\mathcal{C}=\{0,\dots,6\}$ denote the coarse label and
$f\in\mathcal{F}=\{0,\dots,51\}$ the fine-grained label. Each fine-grained class has a unique
parent $c(f)$, and the parents partition $\mathcal{F}$ into groups
$\mathcal{G}_c=\{f:\,c(f)=c\}$. We share a single InternVideo2.5 backbone
$\Phi(\cdot)$ and attach two heads on top of the pooled feature $z=\Phi(x)$:
a coarse head producing $P_{\theta_c}(c\mid x)$ over the seven groups, and a
fine-grained head implemented as a group-conditional softmax that normalizes within
each parent group and outputs $P_{\theta_f}\big(f\mid c(f),x\big)$.

Following the tree structure, we recover the fine distribution by the chain
rule and take the prediction over the full label set:
\begin{equation*}
\label{eq:soft_fusion}
P(f\mid x)=P\big(c(f)\mid x\big)\,P\big(f\mid c(f),x\big),
\qquad
\hat{f}=\arg\max_{f\in\mathcal{F}}P(f\mid x).
\end{equation*}
We term this combination \emph{soft fusion}: the search ranges over all of
$\mathcal{F}$ rather than being restricted to a single predicted group, so a
fine class may be selected even when it lies outside the coarse top-1.

When the fine-grained head is instead a flat $52$-way softmax $P_{\theta_f}(f\mid x)$,
we fuse the two heads as a log-linear product of experts,
\begin{equation*}
\label{eq:poe}
s(f)=\log P_{\theta_f}(f\mid x)+\lambda\,\log P_{\theta_c}\big(c(f)\mid x\big),
\qquad \hat{f}=\arg\max_{f}s(f),
\end{equation*}
where $\lambda\in[0,1]$ weights the coarse signal and is selected on the
validation set by macro-F1.

The two heads are trained jointly with
\begin{equation*}
\label{eq:loss}
\mathcal{L}=\mathcal{L}_{\text{fine}}+\alpha\,\mathcal{L}_{\text{coarse}},
\end{equation*}
where $\mathcal{L}_{\text{coarse}}$ is the cross-entropy on
$P_{\theta_c}(c\mid x)$ and $\mathcal{L}_{\text{fine}}$ is the cross-entropy
on the group-conditional fine-grained head, paired with class-balanced reweighting so
that rare fine-grained classes are not dominated within their group. The weight
$\alpha$ is tuned on the validation set.

When a coarse--fine pair is required as output, we report the
hierarchy-consistent pair $\big(c(\hat{f}),\hat{f}\big)$ induced by the
selected fine-grained label.

\subsection{Lightweight Candidate-Label Reranker}

Fine-grained micro-action recognition is challenging because different categories can correspond to very similar motion patterns. A standard 52-class model may assign comparable probabilities to several related actions, especially when the visual differences between them are subtle. In such cases, the top-1 prediction can be unstable. Repeating a full classification over all 52 categories would add unnecessary computation and would still not explicitly focus on the labels that are most likely to be confused. We address this problem with a lightweight candidate-label reranker. The module does not replace the base action model and does not perform another full 52-class prediction. Instead, it reranks only the candidate labels produced by the fine-tuned model and selects the most likely action category from this restricted set.

To obtain reliable hard samples for training the reranker, we do not directly use the incorrect predictions generated by the fine-tuned model on the training set. These predictions may be affected by sample memorization, because the model has already seen the same samples during optimization. The resulting errors can therefore be biased and may not match the error distribution at test time. We adopt an out-of-fold prediction strategy to reduce this bias. Specifically, the training set is divided into \(F\) folds. In each round, the fine-tuned model is trained on \(F-1\) folds and evaluated on the remaining fold. By rotating the validation fold, each training sample receives a prediction from a model that has not used this sample for training. We record the top-1 prediction, the top-\(K\) predictions, logits, prediction probabilities, and prediction correctness. Compared with direct inference on the training set, out-of-fold prediction provides a more realistic source of hard samples and reduces the bias caused by overfitting to the training data.

Based on the out-of-fold predictions, we construct a candidate label set for each sample. The candidate set always includes the ground-truth action label and several hard negative labels that are likely to be confused with it. These hard negatives are drawn from three sources. The first source is the high-scoring but incorrect top-\(K\) predictions produced by the fine-tuned model. The second source consists of action labels from the same body-motion group as the ground-truth label. The third source consists of frequently confused labels estimated from the validation-set confusion matrix. For example, for a sample with the ground-truth label \texttt{shaking head}, if the fine-tuned model often predicts \textit{turning head} or \textit{nodding}, the candidate set can be constructed as \{\textit{shaking head}, \allowbreak \textit{turning head}, \allowbreak \textit{nodding}, \allowbreak \textit{tilting head}, \allowbreak \textit{head up}\}. This construction encourages the reranker to learn fine-grained differences among similar actions, rather than relying on random negative labels that are already easy to separate.

The training data for the reranker are not restricted to samples that are incorrectly predicted by the fine-tuned model. Although incorrect samples are useful for learning how to correct errors made by the base model, using only such samples may introduce a correction bias. The model may then learn to override the prediction of the base model even when the original prediction is correct. To reduce this risk, the training data include both incorrect samples and a subset of low-confidence correct samples. Incorrect samples provide supervision for error correction, whereas low-confidence correct samples help the reranker learn when the original prediction should be preserved. This design better matches the inference setting, where both wrong predictions and uncertain but correct predictions may occur.

The reranker is implemented as a lightweight video-label matching network. For each video, we first extract a global video feature \(\mathbf{v}\) from the fine-tuned video encoder. For each candidate action label \(c_i\), a learnable label embedding matrix is used to obtain the label representation \(\mathbf{t}_i\). Since MA-52 has a fixed label space, learnable label embeddings can capture implicit relations among action categories in the dataset. This design is suitable for fine-grained reranking, because the model only needs to compare a small set of candidate labels that are likely to be confused with each other.

Given the video feature \(\mathbf{v}\) and the candidate label representation \(\mathbf{t}_i\), we first project them into the same feature dimension. We then construct an interaction feature by concatenating the projected video feature, the projected label feature, their element-wise product, and their absolute difference:
\begin{equation*}
\mathbf{z}_i = [\mathbf{v}, \mathbf{t}_i, \mathbf{v} \odot \mathbf{t}_i, |\mathbf{v} - \mathbf{t}_i|].
\end{equation*}

The element-wise product models the compatibility between the video and the candidate label, whereas the absolute difference captures the discrepancy between their features. The interaction feature \(\mathbf{z}_i\) is fed into a lightweight MLP scorer, which outputs a matching score \(s_i\) for the video-label pair \((\mathbf{v}, c_i)\). For a candidate set containing \(M\) labels, the reranker produces \(M\) matching scores and ranks all candidate labels within the set.

The model is optimized using a candidate-level cross-entropy loss. For each training sample, the supervisory signal is defined as the relative position of the ground-truth label within the candidate set, rather than its original index in the 52-class label space. For instance, given the candidate set \{\textit{}{shaking head}, \textit{turning head}, \textit{nodding}\}, if the ground-truth label \textit{shaking head} is ranked first, the corresponding target index is set to 0. The loss is defined as
\begin{equation*}
\mathcal{L}_{\text{rank}} = - \log \frac{\exp(s_y)}{\sum_{i=1}^{M}\exp(s_i)}.
\end{equation*}
where \(s_y\) denotes the matching score of the ground-truth candidate. This objective directly optimizes the relative ranking within the candidate set and encourages the score of the ground-truth label to be higher than the scores of other candidate labels.

During inference, the base model first predicts the probability distribution over the 52 action classes for an input video. The top-\(K\) classes are selected as candidate labels. For high-confidence samples, the top-1 prediction of the base model is directly retained as the final prediction. For low-confidence samples, or for samples where the probability gap between the top-1 and top-2 predictions is small, the reranker is used to refine the decision. It computes the matching score for each top-\(K\) candidate label and selects the candidate with the highest score as the final action prediction. In this way, the proposed module improves discrimination among confusing fine-grained actions while keeping the additional computation limited to ambiguous samples.

\section{Experiments}
\subsection{Datasets and Metrics}
The MA-52 dataset is a recent benchmark for micro-action recognition and provides a data basis for analyzing subtle human behaviors. It contains 22,422 video samples from 205 subjects, with a total duration of 12.29 hours. In the Micro-Action Analysis Grand Challenge, the final rankings are determined using a subset of 1,138 samples drawn from the MA-52 test set, which contains 5,586 samples in total.

Because sample counts across micro-action categories in MA-52 follow a long-tailed distribution, this study uses $F1_{micro}$ and $F1_{macro}$ to evaluate model performance. $F1_{micro}$ measures overall recognition performance at the sample level, whereas $F1_{macro}$ measures average recognition performance across categories. Using both metrics reduces the bias of $F1_{micro}$ toward frequent categories and gives a more reliable evaluation of categories with limited samples. To provide a unified assessment of coarse-grained action recognition and fine-grained micro-action recognition, this paper uses $F1_{mean}$ as the final evaluation metric.

\begin{equation*}
F1_{mean}
={%
   (F1_{\textit{macro}}^{\textit{body}}
   +F1_{\textit{micro}}^{\textit{body}}
   +F1_{\textit{macro}}^{\textit{action}}
   +F1_{\textit{micro}}^{\textit{action}})%
}/{4}
\end{equation*}

\begin{table}
  \caption{Comparison of F1 scores on the 1,138-samples MAC 2026 test subset.}
  \label{tab:sota}
  \begin{tabular}{c|cc|cc|c}
    \toprule
    Method & $\mathrm{F1}_{\textit{macro}}^{\textit{body}}$ & $\mathrm{F1}_{\textit{micro}}^{\textit{body}}$ & $\mathrm{F1}_{\textit{macro}}^{\textit{action}}$ &
    $\mathrm{F1}_{\textit{micro}}^{\textit{action}}$ & $\mathrm{F1_{mean}}$ \\
    \midrule
    VideoSwinT\cite{liu2022video} & 70.13 & 74.16 & 48.14 & 56.15 & 62.14\\
    Li et al.\cite{li2025progressive} & 83.49 & 86.38 & 64.47 & 71.79 & 76.54\\
    Wang et al.\cite{wang2025combatting} & 81.47 & 87.13 & 64.46 & 74.85 & 76.98\\
    STIA\cite{yu2024end} & 81.74 & 85.50 & 60.74 & 69.42 & 74.35\\
    HiMEAR\cite{xiazhichao} & 84.06 & 86.38 & 68.31 & 72.23 & 77.75\\
    \textbf{Our Method} & \textbf{85.90}  & \textbf{87.96} & \textbf{71.57} & \textbf{74.52} & \textbf{79.99}\\
  \bottomrule
\end{tabular}
\end{table}

\subsection{Implementation Details}
We implement our method in PyTorch 2.1 with CUDA 12.1. For backbone fine-tuning, we choose the learning rate by grid search over 3e-6 to 3e-7 to keep optimization stable. We use mixed-precision training and activation checkpointing by default to reduce memory cost during training. For each training video, we center-crop each frame and resize it to $224 \times 224$ pixels before feeding it into the model. All experiments are run on 8 A100 GPUs.

\begin{table}
\caption{The ablation of fusion strategies on the MA-52 validation dataset}
\label{tab:soft-fusion-ablation}
\begin{tabular}{c|cc|cc|c}
\toprule
Strategy & $\mathrm{F1}_{\textit{macro}}^{\textit{body}}$ & $\mathrm{F1}_{\textit{micro}}^{\textit{body}}$ & $\mathrm{F1}_{\textit{macro}}^{\textit{action}}$ & $\mathrm{F1}_{\textit{micro}}^{\textit{action}}$ & $\mathrm{F1_{mean}}$ \\
\midrule
Fine head only       & 81.54 & 85.22 & 65.03 & 69.22 & 75.25 \\
Hard Cascade         & 84.72 & 86.70 & 67.25 & 70.15 & 77.20 \\
Product-of-Experts   & 83.12 & 86.62 & 68.23 & 71.22 & 77.30 \\
\textbf{Soft Fusion} & \textbf{85.60} & \textbf{87.99} & \textbf{69.50} & \textbf{73.29} & \textbf{79.09} \\
\bottomrule
\end{tabular}
\end{table}

\begin{table}
\caption{The ablation of data processing and reranker on the validation MA-52 dataset}
\label{tab:ablation}
\begin{tabular}{c|cc|cc|c}
\toprule
Method & $\mathrm{F1}_{\textit{macro}}^{\textit{body}}$ & $\mathrm{F1}_{\textit{micro}}^{\textit{body}}$ & $\mathrm{F1}_{\textit{macro}}^{\textit{action}}$ &
$\mathrm{F1}_{\textit{micro}}^{\textit{action}}$ & $\mathrm{F1_{mean}}$ \\
\midrule
Fine-tuning     & 85.60 & 87.99 & 69.50 & 73.29 & 79.09 \\
+ data process  & 85.55 & 88.71 & 69.51 & 75.18 & 79.74 \\
+ Reranker      & 86.84 & 88.71 & 71.51 & 74.67 & 80.44 \\
\textbf{+ Both} & \textbf{87.71} & \textbf{89.47} & \textbf{72.48} & \textbf{75.50} & \textbf{81.29} \\
\bottomrule
\end{tabular}
\end{table}

\subsection{Comparison with State-of-the-Art Methods}

We report our final results on the MA-52 test set (1,138 samples) and provide a comprehensive comparison with representative micro-action recognition approaches. As shown in Table~\ref{tab:sota}, our method achieves the best body-level, action-level, and overall performance, consistently outperforming existing approaches. The framework integrates end-to-end backbone fine-tuning, imbalance-aware training, hierarchical soft fusion, and selective candidate-label reranking. End-to-end fine-tuning adapts video representations to subtle spatiotemporal variations, while imbalance-aware strategies improve recognition reliability under long-tailed class distributions. Hierarchical soft fusion exploits the relationship between body-motion groups and fine-grained actions, whereas selective reranking resolves residual ambiguities among visually similar labels. Together, these components address representation adaptation, label imbalance, hierarchical consistency, and local class confusion within a unified framework. These results validate the integration of adaptive video representation learning, hierarchy-consistent prediction, and targeted ambiguity correction for fine-grained micro-action recognition.

\subsection{Ablation Analysis}

\textbf{Effect of hierarchical fusion strategies.}
Table~\ref{tab:soft-fusion-ablation} compares strategies for exploiting the hierarchy between body-motion groups and fine-grained actions. All hierarchy-aware strategies outperform the fine-grained-head-only baseline, confirming that body-level supervision provides effective structural guidance. Hard cascade restricts predictions to one coarse group and may propagate coarse-level errors. Product-of-Experts avoids this constraint by combining coarse- and fine-level predictions but does not explicitly model their group-conditional dependency. Soft fusion performs best, increasing $\mathrm{F1_{mean}}$ from 75.25 to 79.09 by combining coarse-group and group-conditional fine-grained probabilities while preserving global competition among action classes. Compared with the fine-grained-head-only baseline, it also improves both body-level and action-level metrics, demonstrating consistent benefits across the two label levels. Soft fusion therefore maintains hierarchical consistency without prematurely excluding plausible actions and regularizes fine-grained classification through body-level predictions.

\textbf{Effect of data preprocessing and the lightweight candidate-label reranker.}
Table~\ref{tab:ablation} evaluates the individual and combined contributions of imbalance-aware data preprocessing and selective candidate-label reranking. Data preprocessing increases $\mathrm{F1_{mean}}$ from 79.09 to 79.74, mainly by improving body-level and action-level micro-F1, while both macro-F1 scores remain nearly unchanged. Specifically, action-level micro-F1 increases from 73.29 to 75.18. This suggests that class-balanced sampling, inverse-frequency reweighting, rare-class augmentation, and hierarchical mixup improve overall prediction reliability but are less effective at resolving residual confusion among fine-grained categories.

The reranker produces a larger gain in action-level macro-F1, increasing it from 69.50 to 71.51, and raises $\mathrm{F1_{mean}}$ to 80.44. This result shows that candidate-based correction is particularly effective for visually similar actions and uncertain predictions. By restricting comparisons to hard negatives, labels within the same body-motion group, and frequently confused categories, it directly targets errors remaining after global classification. Combining both components achieves the best performance across all metrics, with an overall $\mathrm{F1_{mean}}$ of 81.29. These complementary gains indicate that data preprocessing improves representation balance and reliability, whereas reranking refines ambiguous fine-grained decisions during inference. Together, they provide a more effective solution than either component alone.

\section{Conclusion}

This paper presents a unified micro-action recognition framework integrating full end-to-end fine-tuning of InternVideo2.5, hierarchical soft fusion, and a lightweight candidate-label reranker. To address the long-tailed label distribution of MA-52, class-balanced sampling and inverse-frequency reweighting are employed to mitigate the dominance of frequent classes during training. InternVideo2.5 is fine-tuned end to end with coarse and group-conditional fine-grained classification heads over a shared video representation, thereby improving coarse-to-fine prediction consistency and reducing errors caused by premature coarse-level decisions. Hierarchical soft fusion preserves global competition among all fine-grained classes while effectively incorporating coarse-group guidance. For ambiguous samples, the reranker leverages hard samples and candidate-level video-label matching to focus on easily confused fine-grained actions during final prediction. Our method achieves a 79.99\% F1-mean on MA-52, surpassing the previous state of the art by 2.24 F1 points and ranking first in the MAR challenge.

\begin{acks}
This work was supported by the Natural Science Foundation of China (62276242), Hefei Municipal Natural Science Foundation (HZR2431), CAAI-MindSpore Open Fund, developed on OpenI Community.

\end{acks}

\bibliographystyle{ACM-Reference-Format}
% \balance  % [arXiv] no-op here: log reported it was called in the second column
\bibliography{references}

\end{document}